\documentclass[10pt,twocolumn]{ICCAS2024}
 
\usepackage{amssymb}
\usepackage{diagbox}
\usepackage{algorithm}
\usepackage{algorithmic}
\usepackage{stfloats}

\usepackage[maxbibnames=40]{biblatex}
\usepackage{caption}
\begin{document}

\title{An Efficient Deep Reinforcement Learning Model for Online 3D Bin Packing\\ Combining Object Rearrangement and Stable Placement}

\author{Peiwen Zhou${}^{*}$, Ziyan Gao${}^{*}$, Chenghao Li, and Nak Young Chong}

\affils{School of Information Science, Japan Advanced Institute of Science and Technology\\
1-1 Asahidai, Nomi, Ishikawa 923-1292, Japan\\
\{s2310071, ziyan-g, chenghao.li, nakyoung\}@jaist.ac.jp}

\thanks{ \noindent
  This work was supported by JSPS KAKENHI Grant Number JP23K03756.\\
  ${}^*$These authors contributed equally to this work.
  }
\abstract{
    This paper presents an efficient deep reinforcement learning (DRL) framework for online 3D bin packing (3D-BPP). The 3D-BPP is an NP-hard problem significant in logistics, warehousing, and transportation, involving the optimal arrangement of objects inside a bin. Traditional heuristic algorithms often fail to address dynamic and physical constraints in real-time scenarios. We introduce a novel DRL framework that integrates a reliable physics heuristic algorithm and object rearrangement and stable placement. Our experiment show that the proposed framework achieves higher space utilization rates effectively minimizing the amount of wasted space with fewer training epochs. 
}


\maketitle


\section{Introduction}
Robotic bin packing has many applications in the fields of logistics, warehousing, and transportation. The 3D Bin Packing Problem (3D-BPP), a well-known NP-hard problems~\cite{i3}, is referred to as an optimization problem of packing multiple objects into a bin(s), while satisfying the bin capacity constraint~\cite{2}. The 3D-BPP can be tackled offline or online depending on whether all objects can be accessible or not. In terms of offline bin packing task, this setting assumes the prior knowledge of all objects, usually, finding the optimal packing sequence and optimal placement are involved in this setting. Typically, meta-heuristic algorithms have been employed to determine the optimal order sequence in previous studies~\cite{4}, thereafter, heuristic algorithms, such as DBLF proposed by Korha and Mustaf~\cite{7} or HM proposed by Wang and Kris~\cite{8}, are leveraged to determine where to place the object into the bin.

\begin{figure}[t]
\begin{center}
\includegraphics[width=7.5cm]{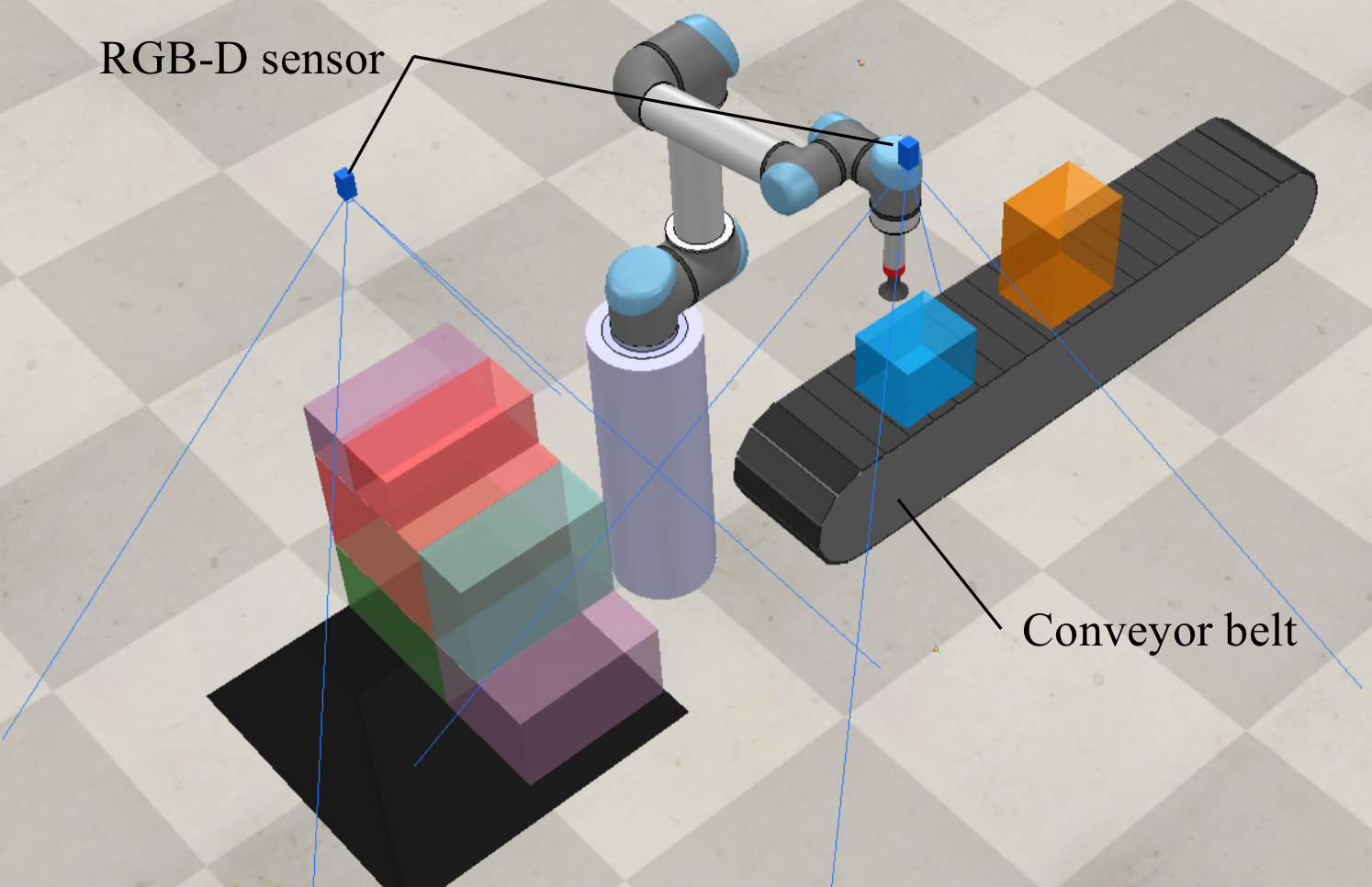}
\caption{Online 3D-BPP, where the agent can only observe an upcoming object and pack it on-the-fly.}\label{robotscene}
\end{center}
\end{figure}

Compared with offline bin packing, online bin packing is more challenging. Basically, the packing order is random, and the agent can only observe the upcoming objects (either single or multiple objects) as illustrated in Fig.~\ref{robotscene}. In this context, relying exclusively on heuristics results in a considerable decline in bin utilization~\cite{4}. Under these constraints, Yang {\it et al.}~\cite{i4} employed unpacking-heuristics to improve the utilization. Nonetheless, this method raises the time cost, thereby diminishing the overall efficiency of the packing process.

Recent progress in DRL has shown promising results in various domains by enabling models to learn optimal policies through trial and error~\cite{19}. Compared with heuristic algorithms, DRL excels in addressing optimization problems effectiveness in complex environments. However, real-world physical law damages the training efficiency as learning the physics in complex environment takes many trial-and-error iterations, and the stable placement cannot be guanranteed. Zhao {\it et al.}~\cite{14} and Yang {\it et al.}~\cite{i4} leveraged neural network to predict the physical feasibility map, enabling the agent to learn feasible packing strategies. Although these methods have achieved promising results in 3D-BPP, object stability is not guaranteed. To address these challenges, we propose an efficient and effective DRL framework using a highly reliable physics heuristic algorithm for online 3D-BPP. The main contributions of this paper are as follows.

\begin{itemize}
    \item We proposed a highly reliable physics heuristic algorithm that guarantees the stability of object placement in complex multi-stack environments, while retaining as many placement positions as possible.
    \item We incorporated an object rearrangement process into the proposed framework which allows the robot manipulator to change the orientation of the upcoming object. It is also an efficient action that directly enhances space utilization without requiring additional time costs.
\end{itemize}

\section{Related Work}

\subsection{Heuristics in Bin Packing Problem}
The bin packing problem is a key challenge in combinatorial optimization, aiming to arrange multiple objects efficiently within the larger container. However, 3D-BPP become unsolvable within a reasonable time frame using exact algorithms~\cite{1} when involving a large number of objects. Over the years, various heuristic and meta-heuristic methods have been developed to address this problem~\cite{5}\cite{6}\cite{7}\cite{9}. Heuristic algorithms critically depend on the sequence of object placement, and current research often employs meta-heuristic techniques such as simulated annealing~\cite{13} and genetic algorithms~\cite{7}.

Consequently, if complete information on all objects to be packed is unavailable, the effectiveness of heuristic algorithms drops significantly. Moreover, in real-world logistics warehouse, gathering detailed information about all objects can be challenging and time-consuming, reducing operational efficiency. Therefore, We propose using the object rearrangement  method to change the orientation of objects in order to improve bin utilization, under the constraints of unchangeable order sequence.

\subsection{DRL in 3D-BPP}
DRL combines the decision-making capabilities of reinforcement learning with the powerful representation learning of deep neural networks. Furthermore, it can be adaptable to changing conditions and provide feasible solutions with highly efficient~\cite{19}, where traditional methods may struggle to find efficient solutions. DRL has recently demonstrated strong performance across various robotics tasks~\cite{17}\cite{18}, showcasing its ability to handle complex spatial and dynamic challenges effectively. 

Thus applying DRL to the 3D-BPP could indeed be a highly efficient approach. For example, Zhao {\it et al.}~\cite{14} introduced a prediction-and-projection scheme where the agent first generates a physical stability mask for placement actions as an auxiliary task, then using this mask to adjust the action probabilities output by the actor during training. However, DRL models can suffer from instability and sensitivity to hyperparameters, making them difficult to tune and sometimes resulting in unpredictable performance. Moreover, most work focuses only on sample constraints, without considering real-world physical properties of objects, including the object CoM and its deviation in a complete stack. These factors can result in solutions that are impractical for real-world applications where physical stability and balance are essential.

Thus we propose the DRL framework integrated with a physics heuristics. This not only guarantees the stability of object placement but also enhances the training efficiency of the model, allowing for faster convergence.

\subsection{Stability check in 3D-BPP}
Stable stacking is a critical factor when designing an online 3D bin packing pipeline. Learning the rules of real-world physics is a very difficult process for DRL. This not only lengthens the training time for the model but also causes fluctuations in model convergence. 

Therefore, for 3D-BPP, it is necessary to design a reliable and efficient physics heuristics for feasible action detection to quickly rule out incorrect actions in the current state. Zhao {\it et al.}~\cite{14} and Yang {\it et al.}~\cite{i4} use the similar scheme that combines the ratio of overlapping parts between the placed item and its contact items with a neural network model for prediction. But this is not a reliable method, since the model is a black box, there are always parts that are inexplicable and unpredictable. On the other hand, Wang and Kris~\cite{8} proposed a mathematical model that using a linear programming method solves for the torque balance and force balance of the object for all contact forces. Although this is a very reliable method,it is too complex for regular objects and usually takes a long time to evaluate all the candidates actions. 

\begin{figure}[t]
\begin{center}
\includegraphics[width=6.5cm]{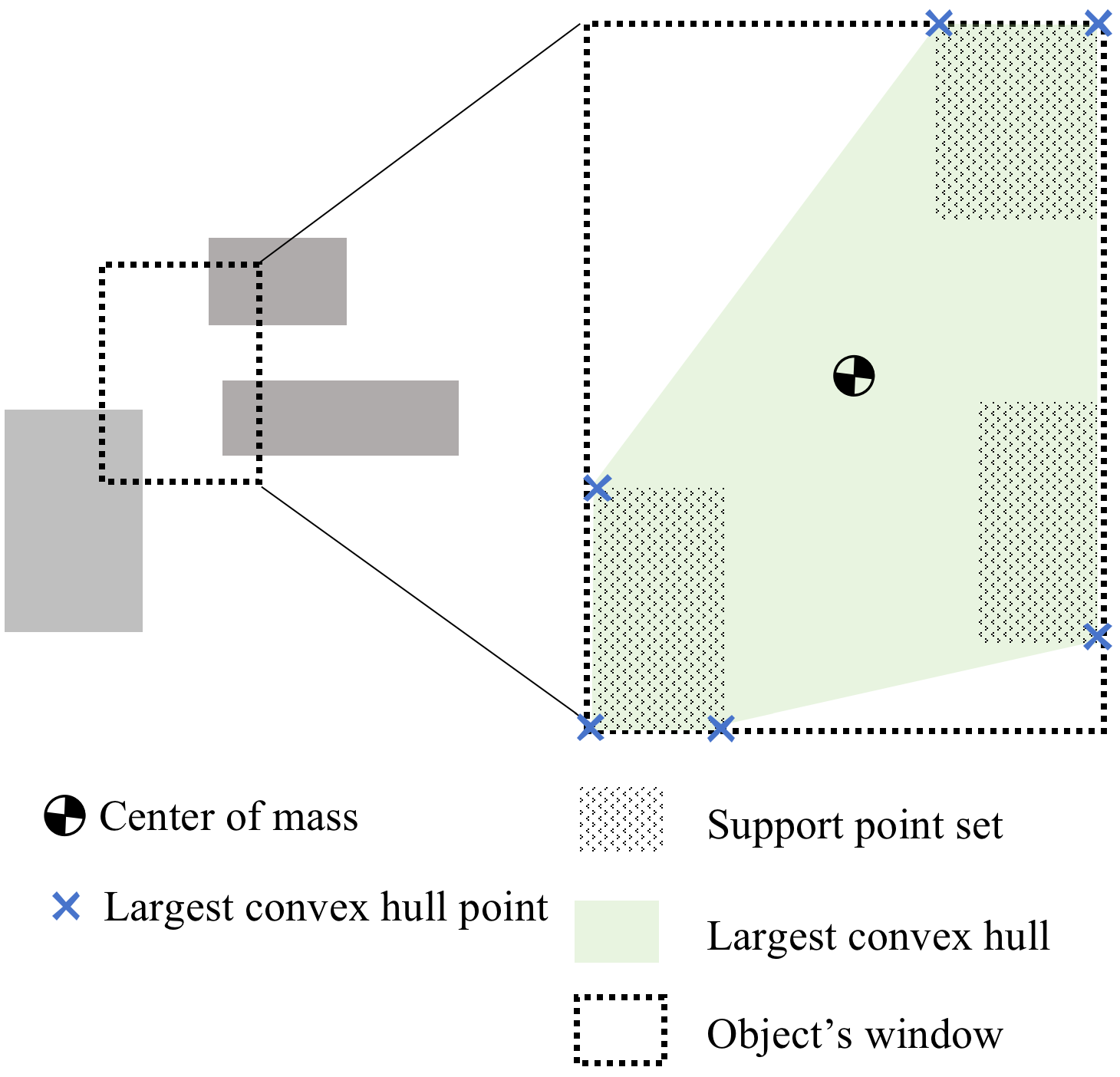}
\caption{The main idea of convexHull-1. The left image depicts a sliding window that matches the size of the incoming object, along with portions of the scene objects contained within the sliding window. The right figure shows the zoom-in version of the content inside the sliding window. To determine the stability of the object, we calculate the largest convex hull of the highest points within the window. Next, we verify whether the center of the window lies within the convex hull. The object is deemed stable when positioned at the center of the sliding window if the convex hull includes the window's center.}\label{windowSliding}
\end{center}
\end{figure}

Thus we propose a new physics heuristic algorithm for rectilinear objects, which can guarantee the stability of object placement in an efficient and effective way, under real-world physical constraint.
\begin{figure*}[t]
\begin{center}
\includegraphics[width=\linewidth]{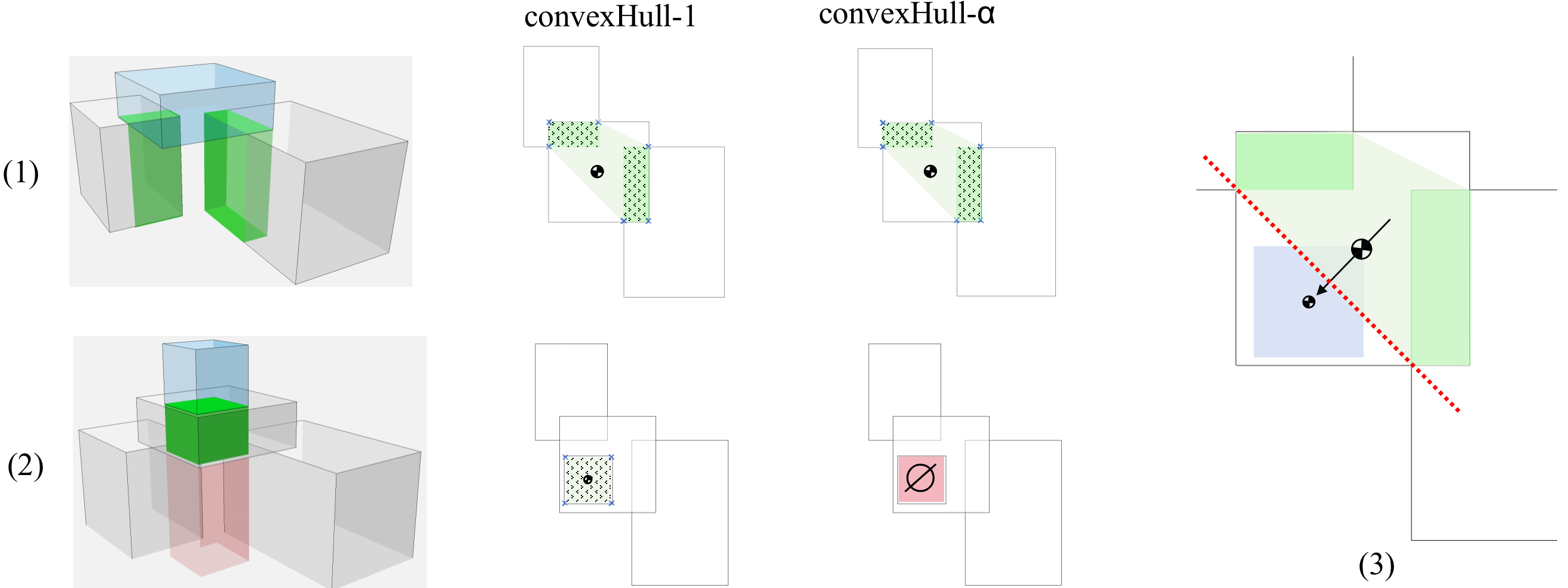}
\caption{Multi-layer packing scenarios showcasing the difference between convexHull-1 and convexHull-$\alpha$ algorithms for checking the stability of the placement. (1) Both convexHull-1 and convexHull-$\alpha$ consider the arrangement to be stable. (2) Conversely, convexHull-1 might incorrectly assess the stability if the incoming object is significantly heavier than the object in the middle layer, as detailed in (3).}\label{StableCheckExample}
\end{center}
\end{figure*}

\section{Method}
We describe our method in two parts. First, we present our stability check method, which is a highlight of our work. Second, we introduce a DRL framework that integrates physical heuristics and object rearrangement.

\subsection{Stability Checking via Physics Heuristics}
In our research, we assume that the object $i\in I$ for bin packing are rigid body and have a uniform mass distribution, so that the center of mass (CoM) is the geometric center of the object. But our method is not limited by the mass distribution, here just for simplify questions. For uneven objects, we can use Gao {\it et al.}~\cite{gao2}\cite{gao1} to estimate as close as possible the CoM.

For the current state bin $(W,D,H)$, we generate a bottom-to-top depth heightmap with a resolution of $0.005$, where each voxel represents a 5 $mm^{3}$ vertical column of 3D space in the agent’s workspace. The object to be placed $i$ is defined by its dimensions $(w_{i},d_{i},h_{i})$. We employ a $w_{i}\times d_{i}$ sliding window to traverse the height map to check the stability of each placement.

Based on physics' principles, we introduce the convexHull-1 method, as shown in Fig.~\ref{windowSliding}. The upward support force, denoted as $p_{f}^{i}= \left\{p_{1},p_{2},...\right\} $ is defined as the set of highest points in the window, obtained by object $i$ under currently placement. We utilize OpenCV~\cite{22} to calculate the largest convex hull formed by $p_{f}$. Then, we evaluate the placement stability by verifying if the center of the sliding window is within the convex hull or not. During our experiments, we observed that relying only on a single layer of the convex hull cannot ensure the stability of object placement. Fig.~\ref{StableCheckExample} shows an example using convexHull-1 for stability check and fail. 

To address the aforementioned issue, we introduce convexHull-$\alpha$, for managing multiple stacks of objects in complex environments. Throughout the object packing procedure, we maintain an empty map with the same size as the action map. The main concept of covexHull-$\alpha$ is that the supporting force must be vertical and originate from the ground. Basically, for each position inside the sliding window, we check the number of wasted voxels along the $z$ axis. We consider that only no wasted voxels can be the reliable support force, which corresponds to the empty map value is zero, denoted as $p_{f}^{\prime}$. After each placement, we update the empty map outlines in Algorithm~\ref{alg2}. Similarly, we use the new set of points $p_{f}^{\prime}$ to calculate the convex hull and determine whether the window's CoM is within it or not. Fig.~\ref{StableCheckExample} illustrates an example of stability check using convexHull-$\alpha$. Algorithm~\ref{alg1} outlines our algorithm in detail.\\
\begin{figure*}[t]
\begin{center}
\includegraphics[width=\linewidth]{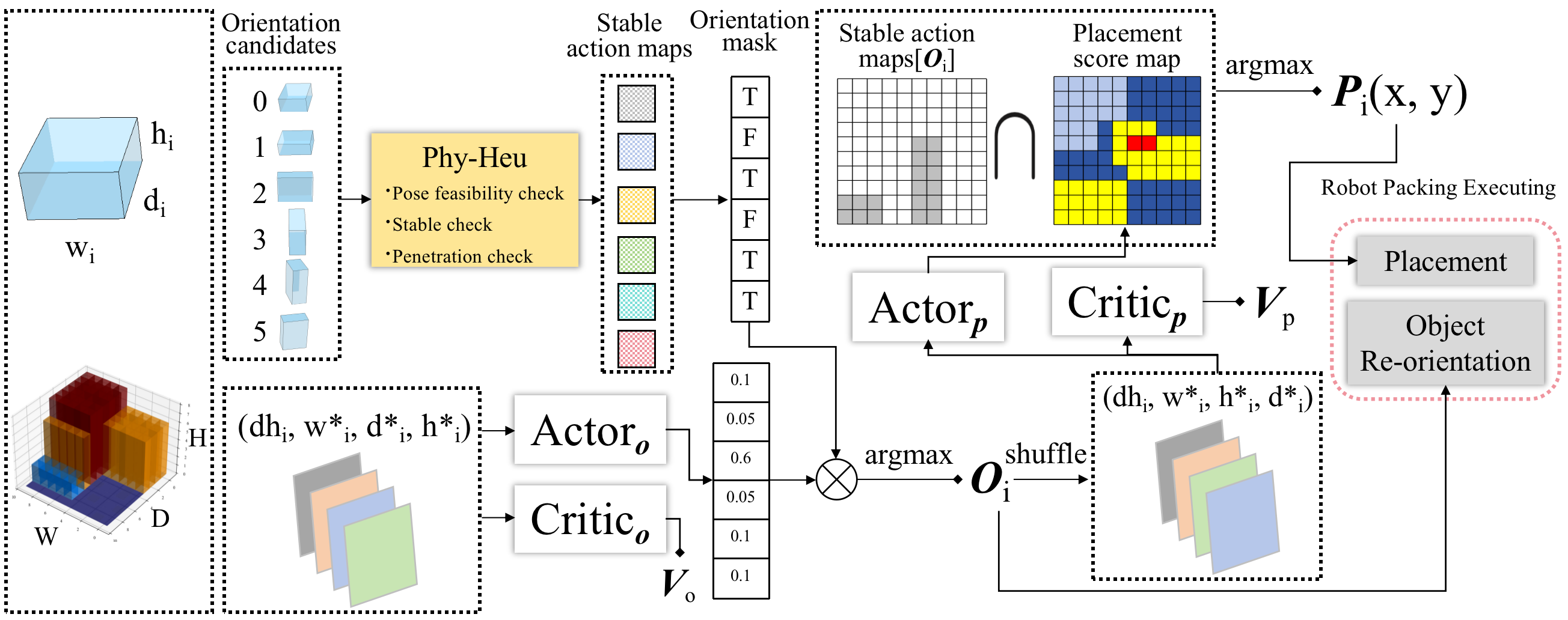}
\caption{The pipeline of the DRL framework combined with object rearrangement and physics heuristics.}\label{model}
\end{center}
\end{figure*}

\begin{algorithm} [thb] 
	\caption{convexHull-$\alpha$} 
	\label{alg1} 
	\begin{algorithmic}
        \REQUIRE Bin depth heightmap $B$, empty map $E$,object$_{i}$ $(w_{i},d_{i},h_{i})$, bin height $H$
        \ENSURE Stable action map $A$
        \FOR{$b_{x,y}\in B$}
            \STATE $window \gets B$[x: x + $w_{i}$][y: y + $d_{i}$]
        \IF{$window$ out of boundary $\OR$ $window.max + h_{i} >H$}
            \STATE$A_{x,y} \gets 0$
            \STATE$continue$
        \ENDIF
        \IF{$window.max$ == 0}
            \STATE$A_{x,y} \gets 1$
            \STATE$continue$
        \ENDIF
        \STATE Find $p_{f}$ in $window$
        \STATE Use $E$ update $p_{f}$ to $p_{f}^{\prime}$
        \STATE $center \gets Point(\frac{w_{i}}{2},\frac{d_{i}}{2})$
        \STATE $hull \gets convexHull(p_{f}^{\prime})$
        \STATE $inside \gets$ Boolean $ center.within(hull) $
        \STATE $A_{x,y} \gets inside$
        \ENDFOR
        \RETURN$A$
    \end{algorithmic} 
\end{algorithm}
\begin{algorithm} [thb]
	\caption{Empty map update} 
	\label{alg2} 
	\begin{algorithmic}
        \REQUIRE Bin depth heightmap $B$, empty map $E$, action point$(X, Y)$, object $i = (w_{i},d_{i},h_{i})$
        \ENSURE Updated empty map $E$
        \STATE$window^{B} \gets B[X:X+w_{i},Y:Y+d_{i}]$
        \STATE$window^{E} \gets E[X:X+w_{i},Y:Y+d_{i}]$
        \STATE$h \gets window^{B}.max$
        \FOR{$e_{x,y}$ in $window^{E}$}
            \STATE $window^{E}_{x,y} \gets e_{x,y} + h - window^{B}_{x,y}$
        \ENDFOR
        \STATE $E[X:X+w_{i},Y:Y+d_{i}] \gets window^{E}$
        \RETURN$E$
    \end{algorithmic} 
\end{algorithm}

\subsection{DRL for Bin Packing}
\subsubsection{Problem Formulation}
Formally, online 3D bin packing involves packing a number of object $i \in I$, each with arbitrary dimensions $(w_{i}, d_{i},h_{i})$ and cuboid shapes, into a bin of arbitrary dimensions $(W,D,H)$. The process is constrained by the visibility of only the immediately upcoming object could be packed into the bin. Once the bin is filled or can not pack upcoming object the process will stop.

To solve this task, we formulate it as a Markov Decision Processes (MDPs), which can be denoted as a tuple $M = \left \langle S,A,P,R,\gamma \right \rangle$. Specifically, we employ two agents with polices $\pi_{o}$ and $\pi_{p}$ to independently predict placement orientation and position.


The whole process is descried as follow: At the time step $t$, the agent observes the environment and takes a state representation, denoted as $s_{t}$. Then the agent $\pi_{o}$ predicts the action $o$ and pass to agent $\pi_{p}$ to predict action $p$. Execute the action tuple$(o_{t},p_{t})$, causing the environment to transition to $s_{t+1}$, then immediately obtains a reward $r_{t}$. The process aims to achieve the maximal cumulative rewards with discount $\gamma$, as shown in Eq.~(\ref{eq.3}) and (\ref{eq.8}), by jointly optimizing two policies.

\begin{align}
&J_{\pi^{*}_{o}} = \max E_{\pi_{o},s_{t},o_{t} = \pi_{o}(s_{t})}[\sum\limits_{t}{\gamma^{t} r_{t}}]\label{eq.3}\\
&J_{\pi^{*}_{p}} = \max E_{\pi_{p},s_{t},p_{t} = \pi_{p}(s_{t}|o_{t})}[\sum\limits_{t}{\gamma^{t} r_{t}}]\label{eq.8}
\end{align}

\subsubsection{State Definition}
We define state $s_{t}$ as the configuration of the bin along with the object that is about to be packed. Use the depth image of the bin to generate a bottom-to-top $W \times D$ depth heightmap $dh_{t}$\cite{14}.
Following the work conducted by Yang {\it et al.}~\cite{i4}, given the object $i$ with dimensions $(w_{i},d_{i},h_{i})$, we create a three channel map with the dimension $W \times D \times 3$. Each channel corresponds to one of the object's dimensions and is fully populated with the respective dimension values. Then combine them as $(dh_{i},w_{i}^{*},d_{i}^{*},h_{i}^{*})$ to represent the State.

\begin{figure}[thb]
\begin{center}
\includegraphics[width=6.5cm]{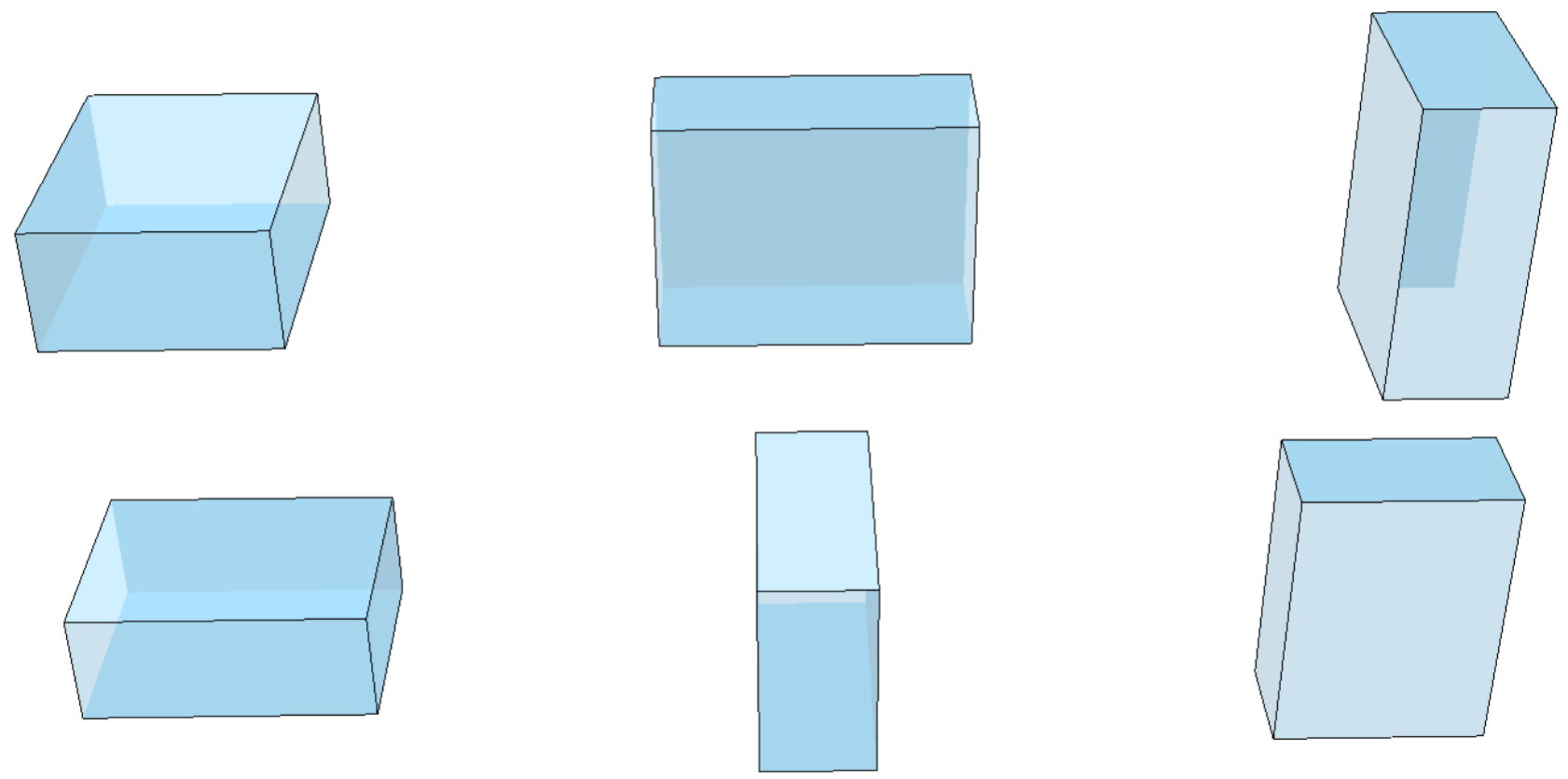}
\caption{Six possible orientations of the packing object.}\label{orientation}
\end{center}
\end{figure}

\subsubsection{Action Definition}
In this work, we propose to arrange object orientation in order to achieve better placement. Therefore, the action is defined as the conjunction of object rearrangement and placement, which is represented by $(o,p)$, where $o$ represents the target object orientation and $p$ represents a specific position on top layer of the bin. To simplify the packing procedure, both $o$ and $p$ are discretized. 

As illustrated in Fig.~\ref{orientation}, there are six different orientations. The number of positions for possible placement is the same as the number of pixels inside the heightmap. Given $(o,p)$, the agent firstly uses object rearrangement operation to achieve the object orientation $o$, and then place the object to the position $p$.

\subsubsection{Reward Function}
Following the idea mentioned in~\cite{14}, at the time step $t$, the immediate reward is the weighted subtraction of increased utilization space and the wasted space given by Eq.~(\ref{eq:5}) to (\ref{eq:7}). Please note that the wasted space can be calculated efficiently by comparing the summation of the empty map before and after the placement. In addition, both $\alpha$ and $\beta$ are set to be one in our experiment. 

\begin{align}
&R^{i} = \alpha \cdot r_{v}^{i} - \beta \cdot r_{waste}^{i} \label{eq:5}\\
&r_{v}^{i} = \frac{w_{i}\cdot d_{i} \cdot h_{i}}{W \cdot D \cdot H} \label{eq:6}\\
&r_{waste}^{i} = \frac{v_{waste}^{i}}{W \cdot D \cdot H}\label{eq:7}
\end{align}

\subsubsection{Physics Heuristics DRL Framework}
Distinct from other works~\cite{i4}, we proposed a two-agents DRL framework integrated with physics heuristics as shown in Fig.~\ref{model}. Based on Proximal Policy Optimization (PPO)~\cite{24}, we develop two actor networks: $Actor_{o}$ dedicated to predicting the object's orientation and $Actor_{p}$ to determining the packing position. Both actor networks takes input as the 4-channels maps, the output of $Actor_{o}$ is a six-dimensional vector where each element dedicates one specific object orientation, the output of $Actor_{p}$ is the action map for placement with the same size as the heightmap. 

The training pipeline is as follows: Given the object $i$ and configuration of bin $dh_{i}$, firstly, the Phy-Heu module generates stable action maps for all potential object orientations. Using these stable action maps, we construct an orientation mask to exclude orientations that do not allow for any feasible stable placement.  Meanwhile, $Actor_{o}$ will predict the probability distribution of the object orientations. Using the orientation mask and the predicted distribution of orientations, the orientation is sampled. Next,  based on the sampled orientation, the agent $Actor_{p}$ takes the $dh_i$ and shuffled $(w^*_i, d^*_i, h^*_i)$ to predict the placement score map. Lastly, we sample the action $P_i$  from the intersected map of the corresponding stable action map and the predicted action score map to ensure the placement stability.

\section{Experiment and Result}
Our experiments were performed with CoppeliaSim using Newton physical engine. The experiments include: (1) the validation of the physical heuristic algorithms; (2) the training and testing of the DRL framework.

\subsection{Physics Heuristics Validation}
We compare the physics heuristic with algorithms convexHull-1 and convexHull-$\alpha$ on CoppeliaSim. The bin dimensions $W = D = H = 0.6 m$. Objects are randomly generated with dimensions $w_{i}, d_{i}, h_{i} \in [0.03,0.3] m $. In this experiment, based on the stable action map computed by convexHull-1 or convexHull-$\alpha$, a random position considered to be stable for placement is selected at each time step. The stability of the bin objects are checked after each placement. The runtime of the two algorithms and the number of un-stable placement are reported in Table \ref{tab1}. Based on Table~\ref{tab1}, We find that convexHull-$\alpha$ significantly surpasses convexHull-1 {\it w.r.t.} the accuracy of stability check. There was only one instance where convexHull-$\alpha$ incorrectly assessed the stability. We suspect this is due to the stable issue of the physical engine. In addition, convexHull-1 and convexHull-$\alpha$ have similar runtime which indicate the efficiency of convexHull-$\alpha$.

\begin{table}[htb]
\caption{Comparison of physics heuristics.}\label{tab1}
\begin{tabu} to \columnwidth {|X[c]|X[c]|X[c]|}\hline
& convexHull-1 & convexHull-$\alpha$ \\\hline
Object number & 3000 & 3000 \\\hline
Fall number & 153 & 1 \\\hline
Time cost(s) & 4203.3  & 4452.6 \\\hline
Per cost(s) & 1.40 - 1.00 & 1.48 - 1.00 \\\hline
Fall rate & 5.1\%  & 0.03\% \\\hline
\end{tabu} 
\end{table}

\subsection{DRL framework result}
RS~\cite{14} bin packing dataset is leveraged to train and test the proposed DRL framework. To evaluate the effectiveness of our proposed method, the result reported in Zhao {\it et al.}~\cite{14} is our baseline as we share the same setting. Consistent with previous studies, we employed space utilization (Uti.) as the metric to evaluate the bin packing policy, where a higher value indicates better performance. We test on the dataset RS, CUT-1, and CUT-2~\cite{14}, which is summarized in Table~\ref{tab2}.

\begin{table}[htb]
\caption{Comparison of packing performance.}\label{tab2}
\begin{center}
\vspace{-0.25cm}
\begin{tabu} to \columnwidth {|X[c]|X[c]|X[c]|X[c]|X[c]|}\hline
& RS & CUT-1 & CUT-2 & epoch \\\hline
Ours & 61.2\% & 63.3\% & 62.5\% & 18.6k \\\hline
\cite{14}& 50.5\% & 60.8\% & 60.9\% & 100k \\\hline
\end{tabu}
\end{center} 
\end{table}

\begin{figure}[thb]
\begin{center}
\includegraphics[width=6cm]{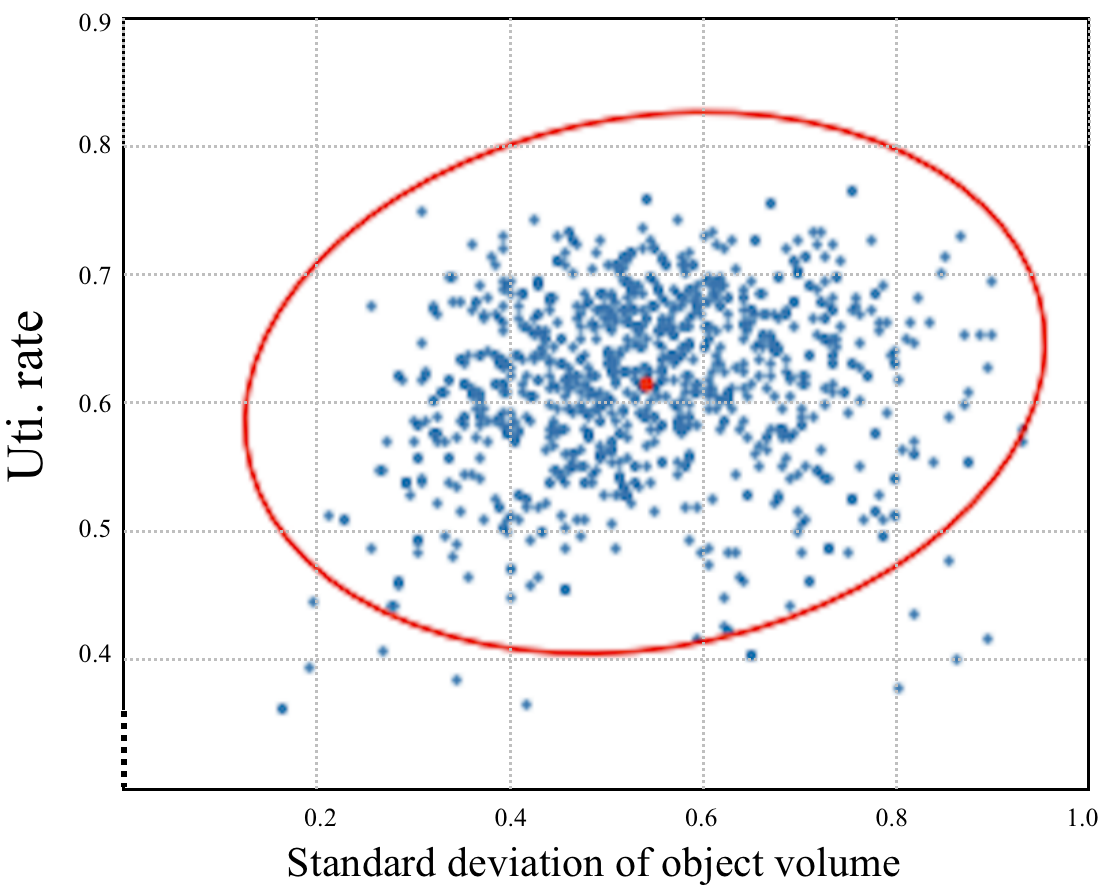}
\caption{Space utilization of our model independent of the standard deviation in object volume.}\label{wUti_std}
\end{center}
\end{figure}

The results show that our method achieves higher Uti. with fewer training epochs. Additionally, we analyzed the RS test results, comparing each test's Uti with the standard deviation of the object volumes in the object sequence. Specifically, larger standard deviation indicates greater volume difference among the objects. As shown in Fig.~\ref{wUti_std}, we found that the our model trained on the RS dataset is not affected by the differences in object volume within the sequence.

\section{Conclusion}
In this paper, we proposed an efficient DRL framework for tackling the online 3D-BPP, integrating a reliable physics heuristic algorithm and object rearrangement technique to enhance the stability and utilization of object placements within bins. Our experiments demonstrated that the proposed method achieves higher utilization rates with fewer training epochs compared to the baseline. The integration of the physics heuristic ensures the stability of object placements, significantly reducing the occurrence of object falls during both training and testing phases. The object rearrangement technique further improves the utilization of bin space without additional time costs, making our framework highly efficient for on-the-fly applications.

In the future, we aim to make our physics heuristic algorithm more accurate by precisely predicting each stable placement position, and to improve the training efficiency of the DRL model. Additionally, we will incorporate the method proposed by Gao {\it et al}~\cite{gao2}\cite{gao1} and Li {\it et al}~\cite{Li} to grasp and pack irregular objects in the real world and attempt to propose a strategy for packing unknown and uneven objects in complex real-world environments.

\printbibliography{}

@article{1,
  title={Exact solution of the two-dimensional finite bin packing problem},
  author={Martello, Silvano and Vigo, Daniele},
  journal={Management Science},
  volume={44},
  number={3},
  pages={388--399},
  year={1998},
  publisher={INFORMS}
}

@article{2,
  title={The three-dimensional bin packing problem},
  author={Martello, Silvano and Pisinger, David and Vigo, Daniele},
  journal={Operations Research},
  volume={48},
  number={2},
  pages={256--267},
  year={2000},
  publisher={INFORMS}
}

@article{3,
  title={Michael r. garey and david s. johnson, computers and intractability: a guide to the theory of np-completeness},
  author={Book, Ronald V},
  year={1980}
}

@article{4,
  title={An empirical investigation of meta-heuristic and heuristic algorithms for a 2D packing problem},
  author={Hopper, EBCH and Turton, Brian CH},
  journal={European Journal of Operational Research},
  volume={128},
  number={1},
  pages={34--57},
  year={2001},
  publisher={Elsevier}
}

@article{5,
  title={Orthogonal packings in two dimensions},
  author={Baker, Brenda S and Coffman, Jr, Edward G and Rivest, Ronald L},
  journal={SIAM Journal on computing},
  volume={9},
  number={4},
  pages={846--855},
  year={1980},
  publisher={SIAM}
}

@article{6,
  title={Worst-case performance bounds for simple one-dimensional packing algorithms},
  author={Johnson, David S. and Demers, Alan and Ullman, Jeffrey D. and Garey, Michael R and Graham, Ronald L.},
  journal={SIAM Journal on computing},
  volume={3},
  number={4},
  pages={299--325},
  year={1974},
  publisher={SIAM}
}

@inproceedings{7,
  title={A hybrid genetic algorithm for packing in 3D with deepest bottom left with fill method},
  author={Karabulut, Korhan and {\.I}nceo{\u{g}}lu, Mustafa Murat},
  booktitle={International Conference on Advances in Information Systems},
  pages={441--450},
  year={2004},
  organization={Springer}
}

@article{8,
  title={Dense robotic packing of irregular and novel 3D objects},
  author={Wang, Fan and Hauser, Kris},
  journal={IEEE Transactions on Robotics},
  volume={38},
  number={2},
  pages={1160--1173},
  year={2021},
  publisher={IEEE}
}

@inproceedings{9,
  title={Two natural heuristics for 3D packing with practical loading constraints},
  author={Wang, Lei and Guo, Songshan and Chen, Shi and Zhu, Wenbin and Lim, Andrew},
  booktitle={PRICAI 2010: Trends in Artificial Intelligence: 11th Pacific Rim International Conference on Artificial Intelligence, Proceedings 11},
  pages={256--267},
  year={2010},
  organization={Springer}
}

@inproceedings{13,
  title={A simulated annealing algorithm for the circles packing problem},
  author={Zhang, Defu and Huang, Wenqi},
  booktitle={International Conference on Computational Science},
  pages={206--214},
  year={2004},
  organization={Springer}
}

@inproceedings{14,
  title={Online 3D bin packing with constrained deep reinforcement learning},
  author={Zhao, Hang and She, Qijin and Zhu, Chenyang and Yang, Yin and Xu, Kai},
  booktitle={AAAI Conference on Artificial Intelligence},
  volume={35},
  number={1},
  pages={741--749},
  year={2021}
}

@article{17,
  title={Grasping in the wild: Learning 6dof closed-loop grasping from low-cost demonstrations},
  author={Song, Shuran and Zeng, Andy and Lee, Johnny and Funkhouser, Thomas},
  journal={IEEE Robotics and Automation Letters},
  volume={5},
  number={3},
  pages={4978--4985},
  year={2020},
  publisher={IEEE}
}

@article{18,
  title={Visual navigation with multiple goals based on deep reinforcement learning},
  author={Rao, Zhenhuan and Wu, Yuechen and Yang, Zifei and Zhang, Wei and Lu, Shijian and Lu, Weizhi and Zha, ZhengJun},
  journal={IEEE Transactions on Neural Networks and Learning Systems},
  volume={32},
  number={12},
  pages={5445--5455},
  year={2021},
  publisher={IEEE}
}

@article{19,
  title={Guided reinforcement learning with learned skills},
  author={Pertsch, Karl and Lee, Youngwoon and Wu, Yue and Lim, Joseph J},
  journal={arXiv preprint arXiv:2107.10253},
  year={2021}
}

@article{22, 
    author = {Bradski, G.}, 
    journal = {Dr. Dobb's Journal of Software Tools}, 
    title = {The OpenCV Library}, 
    year = {2000} 
}

@article{24,
  title={Proximal policy optimization algorithms},
  author={Schulman, John and Wolski, Filip and Dhariwal, Prafulla and Radford, Alec and Klimov, Oleg},
  journal={arXiv preprint arXiv:1707.06347},
  year={2017}
}

@article{i3,
  title={Optimal packing and covering in the plane are NP-complete},
  author={Fowler, Robert J and Paterson, Michael S and Tanimoto, Steven L},
  journal={Information Processing Letters},
  volume={12},
  number={3},
  pages={133--137},
  year={1981},
  publisher={Elsevier}
}

@article{i4,
  title={Heuristics integrated deep reinforcement learning for online 3d bin packing},
  author={Yang, Shuo and Song, Shuai and Chu, Shilei and Song, Ran and Cheng, Jiyu and Li, Yibin and Zhang, Wei},
  journal={IEEE Transactions on Automation Science and Engineering},
  year={2023},
  publisher={IEEE}
}

@inproceedings{Li,
  title={Safety-optimized Strategy for Grasp Detection in High-clutter Scenarios},
  author={Li, Chenghao and Zhou, Peiwen and Chong, Nak Young},
  booktitle={2024 21st International Conference on Ubiquitous Robots (UR)},
  pages={192--197},
  year={2024},
  organization={IEEE}
}

@article{gao1,
  title={Zero moment two edge pushing of novel objects with center of mass estimation},
  author={Gao, Ziyan and Elibol, Armagan and Chong, Nak Young},
  journal={IEEE Transactions on Automation Science and Engineering},
  volume={20},
  number={3},
  pages={1487--1499},
  year={2023},
  publisher={IEEE}
}

@article{gao2,
  title={On the generality and application of Mason's voting theorem to center of mass estimation for pure translational motion},
  author={Gao, Ziyan and Elibol, Armagan and Chong, Nak Young},
  journal={IEEE Transactions on Robotics},
  volume={40},
  pages={2656--2671},
  year={2024},
  publisher={IEEE}
}
\end{document}